\documentclass[10pt,twocolumn,letterpaper]{article}

\usepackage{times}
\usepackage{epsfig}
\usepackage{graphicx}
\usepackage{amsmath}
\usepackage{amssymb}
\usepackage{xcolor}
\usepackage{enumitem}
\usepackage{flushend}
\setlist{nolistsep}
\usepackage{authblk}

\renewcommand\paragraph[1]{\vspace{1mm} \noindent  \textbf{#1} }

\definecolor{gold}{HTML}{7f851f}
\definecolor{greenish}{HTML}{418a3c}

\begin{document}

\title{Depth from Videos in the Wild:\\Unsupervised Monocular Depth Learning from Unknown Cameras}

\author[1,2]{Ariel Gordon}
\author[1]{Hanhan Li}
\author[1,2]{\vspace{-2mm} Rico Jonschkowski}
\author[1,2]{Anelia Angelova}
\affil[ ]{\tt\small \{\vspace{1mm}gariel,uniqueness,rjon,anelia\}@google.com}
\makeatletter
\renewcommand\AB@affilsepx{, \protect\Affilfont}
\makeatother

\date{}
\affil[1]{Google AI}
\affil[2]{Robotics at Google \vspace{-1mm}}


\maketitle

\begin{abstract} \emph{
   We present a novel method for simultaneous learning of depth, egomotion, object motion, and camera intrinsics from monocular videos, using only consistency across neighboring video frames as supervision signal.
   Similarly to prior work, our method learns by applying differentiable warping to frames and comparing the result to adjacent ones, but it provides several improvements: We address occlusions geometrically and differentiably, directly using the depth maps as predicted during training. We introduce randomized layer normalization, a novel powerful regularizer, and we account for object motion relative to the scene. 
   To the best of our knowledge, our work is the first to learn the camera intrinsic parameters, including lens distortion, from video in an unsupervised manner, thereby allowing us to extract accurate depth and motion from arbitrary videos of unknown origin at scale. 
   We evaluate our results on the Cityscapes, KITTI and EuRoC datasets, establishing new state of the art on depth prediction and odometry, and demonstrate qualitatively that depth prediction can be learned from a collection of YouTube videos.}
\end{abstract}

\section{Introduction}

\begin{figure}[ht!]
\begin{center}
   \includegraphics[width=0.49\linewidth]{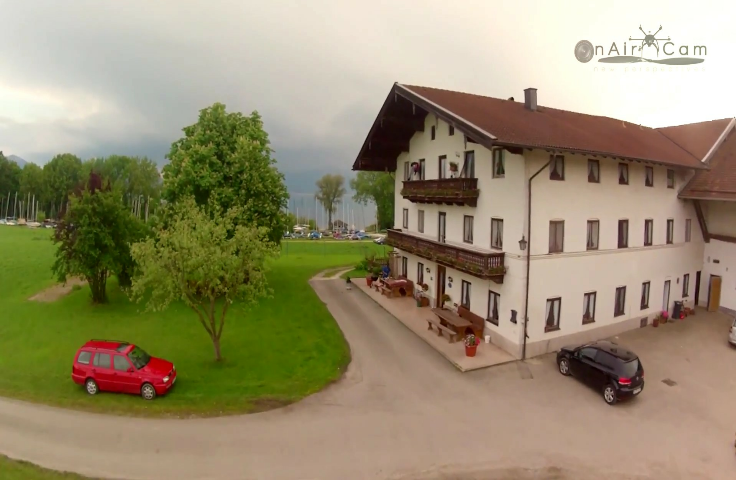}
\includegraphics[width=0.49\linewidth]{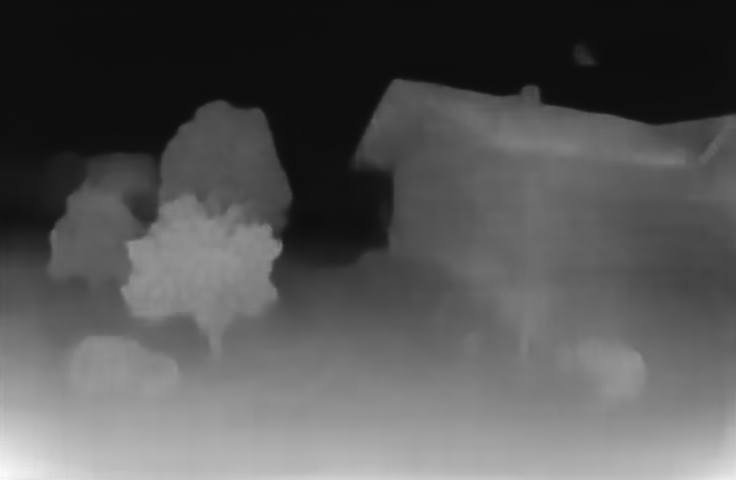}
\includegraphics[width=0.49\linewidth]{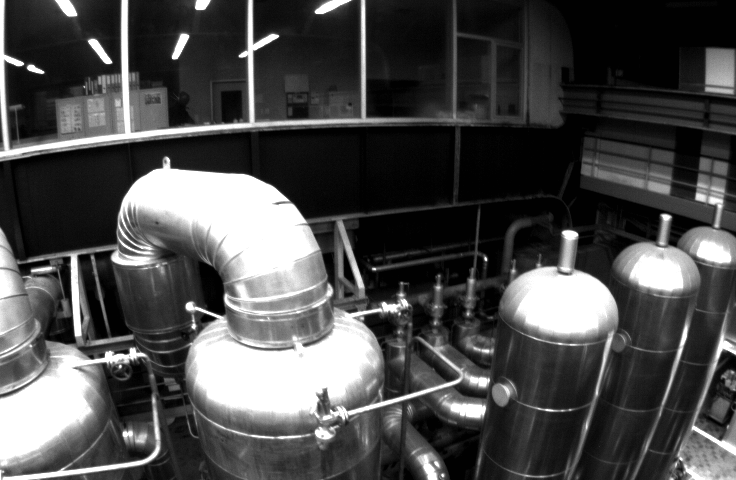}
\includegraphics[width=0.49\linewidth]{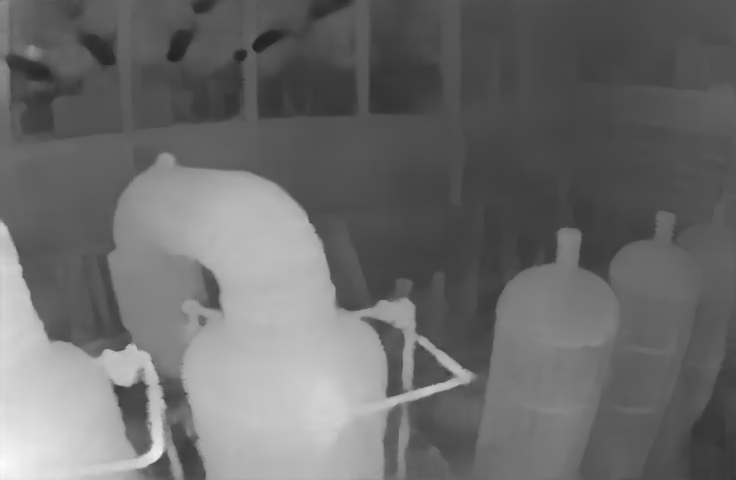}
\includegraphics[width=0.49\linewidth]{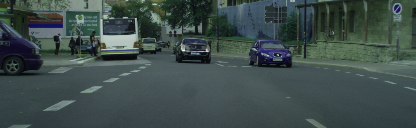}
\includegraphics[width=0.49\linewidth]{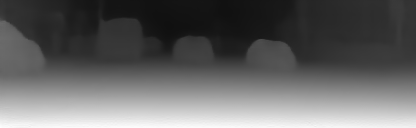}
\includegraphics[width=0.49\linewidth]{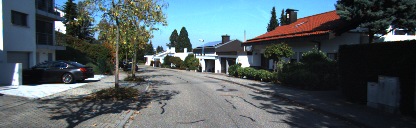}
\includegraphics[width=0.49\linewidth]{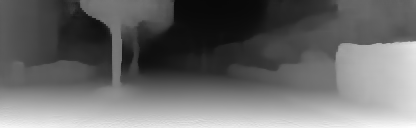}

\end{center}
   \caption{\small Qualitative results of our approach for learning depth from videos of unknown sources, which is enabled by simultaneously learning the camera extrinsic \emph{and intrinsic} parameters. Since our method does not require knowing the camera parameters, it can be applied to any set of videos. All depth maps (visualized on the right, as disparity) were learned from raw videos without using the any camera intrinsics groundtruth.
   From top to bottom: frames from YouTube8M \cite{YouTube8M}, from EuRoC MAV dataset \cite{Burri2016Euroc}, from Cityscapes \cite{Cordts2016Cityscapes} and from KITTI. \vspace{-1mm}}
\label{fig:intro}
\end{figure}

Estimating 3D structure and camera motion from video is a key problem in computer vision. Traditional approaches to this problem rely on identifying the same points in the scene in multiple consecutive frames and solving for a 3D structure and camera motion that is maximally consistent across those frames \cite{sift}. But such correspondences between frames can only be established for a subsets of all pixels, which leaves the problem of estimating depth underdetermined. As commonly done with inverse problems, the gaps are filled with assumptions such as continuity and planarity.

Rather than specifying these assumptions manually, deep learning is able to obtain them from data.
Wherever information is insufficient to resolve ambiguities, deep networks can produce depth maps and flow fields by generalizing from prior examples they have seen. Unsupervised approaches allow learning from raw videos alone, using similar consistency losses as traditional methods but optimizing them during training. At inference, the trained networks are able to predict depth from a single image and egomotion from pairs or longer sequences of images.

As research in this direction got traction \cite{zhou2017unsupervised,garg2016unsupervised,godard2017monodepth,ummenhofer2017demon,mahjourian2018unsupervised,vijayanarasimhan2017sfmnet}, it became clear that object motion is a major obstacle because it violates the assumption that scene is static. Several directions have been proposed to address the issue \cite{yin2018geonet,yang2018every}, including leveraging semantic understanding of the scene through instance segmentation \cite{casser2019struct2depth}. Occlusions have been another limiting factor, and lastly, in all prior work in this direction, the intrinsic parameters of the camera had to be given.

This work addresses these problems and, as a result, reduces supervision and improves the quality of depth and motion prediction from unlabeled videos. First, we show that a deep network can be trained to predict the \emph{intrinsic} parameters of the camera, including lens distortion, in an \emph{unsupervised} manner, from the video itself (see Fig.~\ref{fig:intro}). Second, we are the first in this context to address occlusions directly, in a geometric way, from the predicted depth as it is. Lastly, we substantially reduce the amount of semantic understanding needed to address moving elements in the scene: Instead of segmenting every instance of a moving object and tracking it across frames \cite{casser2019struct2depth}, we need a single mask that covers pixels that \emph{could} belong to a moving object. This mask can be very rough, and in fact can be a union of rectangular bounding boxes. Obtaining such a rough mask is a much simpler problem and can be solved more reliably with existing models than instance segmentation.

In addition to these qualitative advances, we conduct an extensive quantitative evaluation of our method and find that it establishes a new state of the art on multiple widely used benchmark datasets. Pooling datasets together, a capability which is greatly advanced by our method, proves to enhance quality. Finally, we demonstrate for the first time that depth and camera intrinsics prediction can be learned on YouTube videos, which were captured with multiple different cameras, each with unknown and generally different intrinsics.

\section{Related work}
Estimating scene depth is an important task for robot navigation and manipulation and historically much research has been devoted to it including large body of research on stereo, multi-view geometry, active sensing and so on
~\cite{Saxena2007depth,liu2015single,eigen2014depth}.
More recently learning-based approaches for depth prediction have taken center stage based on learning of dense prediction~\cite{eigen2014depth,liu2015learning,laina2016deeper,cao2016estimating}. In these, scene depth is predicted from input RGB images and the depth estimation function is learned using supervision provided by a sensor, such as a LiDAR. Similar approach is used for other dense predictions e.g. surface normals~\cite{eigen2015normals,Wang2015Designing}. 

\paragraph{Unsupervised depth learning.}
Unsupervised learning of depth, where the only supervision is obtained from the monocular video itself and no depth sensors are needed, has also been popularized recently~\cite{zhou2017unsupervised,garg2016unsupervised,godard2017monodepth,ummenhofer2017demon,mahjourian2018unsupervised,vijayanarasimhan2017sfmnet,yin2018geonet}.
Garg et al.~\cite{garg2016unsupervised} introduced joint learning of depth and ego-motion. Zhou et al.~\cite{zhou2017unsupervised} demonstrated a fully differentiable such approach where depth and ego-motion are predicted jointly by deep neural networks. Techniques were developed for the monocular~\cite{ummenhofer2017demon,Yang2017unsupervised,mahjourian2018unsupervised,yang2018lego,yin2018geonet,wang2018learning,casser2019struct2depth} and binocular~\cite{godard2017monodepth,ummenhofer2017demon,yang2018every,zhan2018unsupervised,wang2018learning,zhan2018unsupervised} settings. It was shown that depth quality at inference is improved when stereo inputs are only used during training, unlike methods that relied on stereo disparity at inference too~\cite{kendall2017end,khamis2018stereonet,yao2018mvsnet}.
Other novel techniques include the use of motion~\cite{yang2018lego,wang2018learning,yin2018geonet,casser2019struct2depth,yang2018every} 

\paragraph{Learning from images or videos in the wild.} Learning depth from images in the wild is also an active research field, mostly focusing on single or multi-view images~\cite{agarwal2009building,Schonberger2016sfm,li2018megadepth}. It is especially hard as shown by Li et al for internet photos, due to the diversity of input sources and no knowledge of the camera parameters~\cite{li2018megadepth}. Our work makes a step in addressing this challenge by learning intrinsics for videos in the wild.

\paragraph{Occlusion aware learning.}
Multiple approaches have been proposed for handling occlusions in the context of optical flow~\cite{wang2018occlusion,janai2018multi,neoral2018continual}.
These approaches are disconnected from geometry.
Differentiable mesh rendering, has recently been attracting increasing attention~\cite{nguyen2018rendernet, kato2018renderer}, begins to adopt a geometric approach to occlusions. In the context of learning to predict depth and egmotion, occlusions were addressed via a learned explainability mask~\cite{zhou2017unsupervised}, by penalizing the minimum reprojection loss between the previous frame or the next frame
into the middle one, and through optical flow \cite{yang2018every}. In the context of unsupervised learning of depth from video, we are the first to propose a direct geometric approach to occlusions via a differentiable loss.

\paragraph{Learning of intrinsics.}
Learning to predict the camera intrinsics has mostly been limited to strongly supervised approaches. The sources of groundtruth varies: Workman et al.~\cite{workman2015deepfocal} use focal lengths estimated employing classical 1D structure from motion. Yan et al. \cite{FocalLens} obtain the focal length based on EXIF. Bogdan et al.~\cite{Bogdan_2018} synthesize images from panoramas using virtual cameras with known intrinsics, including distortion. To our knowledge, our approach is the only one that learns the camera intrinsics in an unsupervised manner, directly from video, jointly with depth, ego-motion and object motion.

\section{Preliminaries}
\label{sec:prelim}
Our method extends prior techniques in the domain of simultaneous learning of depth and motion \cite{Zhou2017,godard2017monodepth,yin2018geonet,tulsiani2017multiview}.
Similarly to prior work, the backbone of our method is the equation that ties together two adjacent video frames using a depth map and the camera matrix: 
\begin{equation}\label{warp_expand}
z'p' = KRK^{-1}zp +  K t
\end{equation}
where K is the intrinsic matrix, namely
\begin{equation}\label{K}
     K = \begin{pmatrix}  
     f_x & 0 & x_0 \\ 0 & f_y & y_0 \\ 0 & 0 & 1 
     \end{pmatrix},
\end{equation}
and $p$ and $p'$ are pixel coordinates in homogeneous form before and after the transformation represented by the rotation matrix $R$ and the translation vector $t$. $z$ and $z'$ are the respective depths, and $f_x, f_y, x_0, y_0$ are the camera intrinsics.

Using $z$, $R$ and $t$ as predicted by deep networks, Eq.~\ref{warp_expand} is used to warp one video frame onto the other. The result is then compared to the actual other frame, and the differences constitute the main component of the training loss. The premise is that through penalizing the differences, the networks will learn to correctly predict $z$, $R$ and $t$.

\section{Method}
In this work we propose simultaneous learning of depth, egomotion, object motion, and camera intrinsics from monocular videos.
To accomplish that, we design a motion-prediction network which predicts camera motion, motion of every pixel with respect to the background, and the camera intrinsics: focal lengths, offsets and distortion.
A second network predicts depth maps. Through imposing consistency across neighboring frames as a loss, the networks simultaneously learn to predict depth maps, motion fields and the camera intrinsics. We introduce a loss that demands consistency only for pixels that are unoccluded in both frames, where occlusion estimation is done geometrically, based on the depth maps themselves as they are learned. The motion fields are regularized with the help of a mask that indicates pixels that might belong to moving objects, obtained from a pretrained segmentation or an object detection network.

\subsection{Learning the intrinsics}
During training, the supervision signal of inter-frame consistency propagates back through $p'$ to the learned quantities $K$, $R$, $z$ and $t$. Since Eq.~\ref{warp_expand} only depends on $K$ via $Kt$ and $K^{-1}RK$, the training loss will drive these two quantities to the correct values, but $Kt$, for example, can be perfectly correct even if $K$ and $t$ are incorrect. In fact, if the network predicted an incorrect intrinsic matrix $\tilde K$, and an incorrect translation vector $\tilde t=\tilde K^{-1}K t$, $\tilde K \tilde t$ still equals $Kt$, so the training loss is not affected.

While translations provide no supervision signal for $K$, fortunately rotations do. The derivation showing that no $\tilde K$ and $\tilde R$ exist such that $\tilde K \tilde R \tilde K^{-1}= K R K^{-1}$ is done in the Appendix. Eq.~\ref{fcond}, derived therein too, ties the tolerance with which the focal lengths can be determined ($\delta f_x$ and $\delta f_y$, denominated in pixels) from two frames, to the amount of camera rotation that occurs between the two:
\begin{equation}\label{fcond}
\delta f_x < \frac{2f_x^2}{w^2 r_y}; \quad \delta f_y < \frac{2f_y^2}{h^2 r_x}.
\end{equation}
$r_y$ and $r_x$ are the rotation angles along the respective axes in radians and $w$ and $h$ are the image width and height respectively.

\subsection{Learning object motion}
\label{sec:objmo}
\begin{figure}[ht]
\begin{center}
   \includegraphics[width=0.96\linewidth]{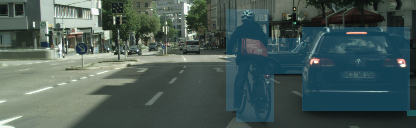}
\includegraphics[width=0.96\linewidth]{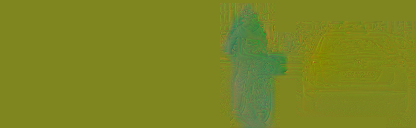}
\end{center}
   \caption{\small Use of a ``possibly mobile" mask to regularize the translation field. An object detection network identifies all instances of objects that are capable of motion, such as pedestrians, cyclists and cars. The union of the bounding boxes comprises the ``possibly mobile" mask, within which the translation field is allowed to vary. The top picture, from Cityscapes, illustrates the mask, and the bottom one is the translation field predicted by the network ($x$, $y$, $z$ coded as RGB). The \textbf{\textcolor{gold}{golden background}} corresponds to motion in the negative $z$ direction, as the entire scene is moving towards the camera. The \textbf{\textcolor{greenish}{greenish silhouette}} is the cyclist moving slightly to the left and slightly towards the camera. Note that the network carves the silhouette out of rough mask.}
\label{fig:boxify}
\end{figure}

Eq.~\ref{warp_expand} can propagate frame inconsistency losses to $z$, $R$ and $t$ at every pixel. However without further regularization, the latter trio remain greatly underdetermined. While continuity of $z$, $R$ and $t$ is a powerful regularizer, we found that further regularization helps significantly. In particular, we impose constancy of $R$ throughout the image, and allow $t$ to deviate from a constant value only at pixels that are designated as ``possibly mobile". This mask can be obtained from a pretrained segmentation model. Unlike in prior work \cite{casser2019struct2depth}, instance segmentation and tracking are not required, as we need a single ``possibly mobile" mask. In fact, we show that a union of bounding boxes is sufficient (see Fig.~\ref{fig:boxify}). In addition, an $L1$ smoothing operator is applied on $t$.

\subsection{Occlusion-aware consistency}
\label{sec:occlusion}
When the camera moves relatively to a scene, and / or objects  move, points in the scene that were visible in one frame may become occluded in another, and vice versa. In pixels that correspond to these points, cross-frame consistency cannot be enforced by a loss. Given a depth map and a motion field in one frame, one could actually detect where occlusion is about to occur, and exclude the occluded areas from the consistency loss.

While detecting occluded pixels is doable, it requires some sort of reasoning about the connectivity of the surface represented by the depth map, and z-buffering. Keeping the mechanism differentiable and efficient, to be suitable for a training loop, may pose a challenge. 

\begin{figure}[ht]
\begin{center}
  \includegraphics[width=0.95\linewidth]{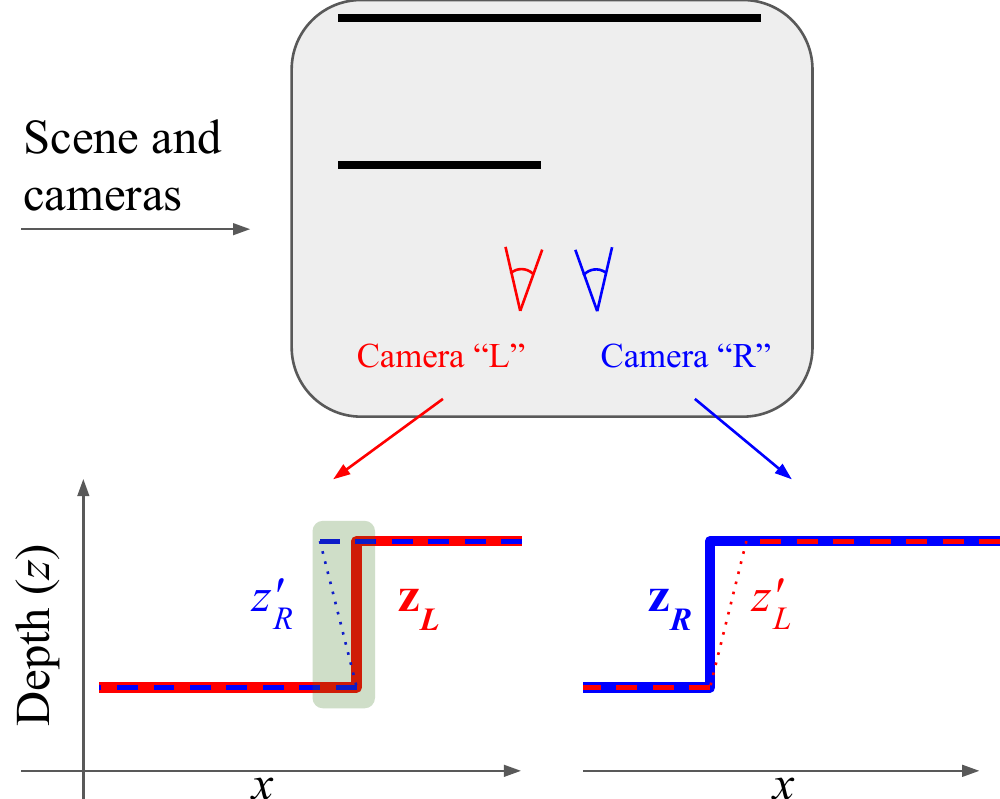}

\end{center}
  \caption{\small An illustration of our proposed method for handling occlusions. At the top we show a two-dimensional ``scene", consisting of two straight surfaces, one partially occluding the other. Two cameras, left (``$L$") and right (``$R$"), are observing the scene. Our method is \emph{monocular}, so these represent two locations of the same camera that moved, and ``left" and ``right" are used for convenience. At the bottom, the depth map observed by each camera is illustrated in as a solid line on the respective side ($\mathbf z _L$ and $\mathbf z _R$). A dashed line shows the depth map obtained from warping one view onto the other ($z'_R$ and $z'_L$). The warped depth map can become a multivalued function, which indicates occlusions (see green shaded rectangle). To handle that, we apply photometric and geometric losses only at pixels where $z'_R\leq\mathbf z _R$ and $z'_L\leq\mathbf z _L$. When the depth maps and motion estimation are correct, the loss in this scheme would indeed evaluate to zero.}
  \label{fig:occdiag}
\end{figure}

We therefore take a different approach, as illustrated in Fig.~\ref{fig:occdiag}. For each pixel $(i, j)$ in the source frame, the predicted depth $z_{ij}$ and the camera intrinsic matrix are used to obtain the respective point in space, $(x_{ij}, y_{ij}, z_{ij})$. The point is moved in space according to the predicted motion field. In particular, the depth changes to $z'$. The new spatial location is reprojected back onto the camera frame, and falls at some generally-different location ($i', j'$) on the target frame. $i'$ and $j'$ are generally non-integer. Therefore obtaining the depth on the target frame at ($i', j'$), $z^t_{i', j'}$, requires interpolation.

Occlusions happen at ($i', j'$) where $z'$ becomes multivalued. At such points, color and depth consistency should be applied only to the visible branch of $z'$, that is, the branch where $z'$ is smaller. If the source and target frames are nearly consistent, the visible branch will be close to target depth at ($i', j'$), $z^t_{i', j'}$. The way we propose to detect that is to include in the losses only points $(i', j')$ where $z'_{i', j'} \leq z^t_{i', j'}$. In other words, only if a transformed pixel on the source frame lands in front of the depth map in the target frame, do we include that pixel in the loss. This scheme is not symmetrical with respect to interchanging the source and target frames, which is why we always apply it in a symmetrized way: We transform the source onto the tagret, calculate the losses, and then switch the roles of source and target. Fig.~\ref{fig:occdiag} illustrates the method. This way of applying losses can be invoked for many types of loss, and we shall refer to it in this paper as ``occlusion-aware" loss or penalty.

\subsection{Networks, losses and regularizations}\label{nets_and_losses}
\paragraph{Networks}
We rely on two convolutional networks, one predicting depth from a single image, and the other predicting egomotion, object motion field relative to the scene, and camera intrinsics from two images. The depth prediction network is a UNet architecture with a ResNet 18 base and a softplus activation ($z=\log(1+\mathrm e^\ell)$) to convert the logits ($\ell$) to depth ($z$).

The motion estimation network is a UNet architecture inspired by FlowNet \cite{flownet}. A stack of convolutions with stride 2 (the``encoder"), with average pooling in the last one, forms a bottleneck of 1024 channels with a 1x1 spatial resolution. Two 1x1 convolutions with 3 output channels each predict the global rotation angles ($r_0$) and the global translation vector ($t_0$). The latter two represent the movement of the entire scene with respect to the camera, due to camera motion. Each of the intrinsic parameters is predicted by a 1x1 convolution stemming from the bottleneck, with softplus activations for the focal lengths and distortions. The next layers progressively refine (that is, increase the spatial resolution of) the translation, from a single vector to a residual translation vector field $\delta t(x, y)$, by a factor of 2 in the height and width dimension each time.

It is here where we utilize the foreground mask $m(x, y)$ described in Sec.~\ref{sec:objmo}: The translation field is expressed as the sum of the global translation vector plus the masked residual translation:
\begin{equation}
t(x, y) = t_0 + m(x, y)\delta t(x, y).
\end{equation}
$m(x, y)$ equals one at pixels that could belong to mobile objects and zero otherwise.

\paragraph{Losses}
Given a pair of frames, we apply an occlusion-aware L1 penalty for each of the RGB color channels and the depths. For the motion fields, we demand cycle consistency: The rotation and translation at pixel $(i, j)$ of the source frame must form the \emph{opposite} transform of the ones at $(i', j')$ of the target frame, and vice versa. Occlusion awareness is invoked here too.

Structural similarity (SSIM) is a crucial component of the training loss, and occlusion awareness as defined above is harder to enforce here, because SSIM involves the neighborhood of each pixel. It is possible that $z' \leq z^t$ holds for only part of the pixels in the neighborhood. The solution we found here is to weigh the structural similarity loss by a function that falls off where the depth discrepancy between the frames is large compared to the RMS of depth discrepancy.

\paragraph{Randomized layer normalization}
Initially our depth prediction network had batch normalization. However we repeatedly observed it leading to anomalous behavior:

\begin{itemize}
\item Eval accuracy was consistently better when running inference at the ``training mode" of batch normalization. That is, instead of long-term averaged means and variances, the ones obtained from the image itself during inference were used\footnote{Even at batch size 1, there would still be means and variances over the spatial dimensions.}, rendering batch normalization more similar to layer normalization \cite{ba2016layer}. 
\item As we increased the batch size at training, the eval accuracy was consistently worse and worse, no matter how we scaled the learning rate.
\end{itemize}

These two observations led us to try replacing batch normalization by layer normalization, but the eval quality degraded. Our next postulate was that while batch normalization is actually acting like layer normalization, the other other items in the batch serve as a source of noise on top of that. To test this theory we replaced batch normalization by layer normalization with Gaussian noise applied on the layer means and variances. 

Indeed all eval metrics exhibited significant improvements compared to batch normalization. Moreover, now the eval metrics started to slightly improve when the batch size at training increased (accompanied with a linear increase of the learning rate \cite{goyal2017accurate}), rather than significantly degrading. The best results were obtained with multiplicative noise. While it has been observed in the past that noise can act as a regularizer, for example with dropout \cite{dropout} or when applied to the gradients \cite{neelakantan2015adding}, we are not aware of prior work where it is applied with layer normalization.

\section{Experiments}

In this section, we evaluate our method on depth prediction, odometry estimation, and the recovery of camera intrinsics across a range of diverse datasets, which will be described next.
\subsection{Datasets}
\label{sec:datasets}
\paragraph{KITTI}
The KITTI dataset is collected in urban environments and
is the main benchmark for depth and ego-motion estimation.
 It is accompanied with a LIDAR sensor which is used for

\paragraph{Cityscapes}
The Cityscapes dataset  is a more recent  
urban driving dataset, which we use for both training and evaluation.
It is a more challenging dataset with many dynamic scenes. With a few exceptions~\cite{pilzer2018unsupervised,casser2019struct2depth} it has not been used for depth estimation evaluation.
We use depth from the disparity data for evaluation on a standard evaluation set of  1250~\cite{pilzer2018unsupervised,casser2019struct2depth}.

\paragraph{EuRoC Micro Aerial Vehicle Dataset}
The EuRoC Micro Aerial Vehicle (MAV) Dataset~\cite{Burri2016Euroc} is a very challenging dataset collected by an aerial vehicle indoors.  
While the data contains a comprehensive suite of sensor measurements, including stereo pairs, IMU, accurate Leica laser tracker ground truth, Vicon scene 3d scans, and camera calibration, we only used the monocular videos for training. Since the camera has significant lens distortion, this is an opportunity to test our method for learning lens distortions (see later sections).

\paragraph{YouTube8M videos}
To demonstrate that depth can be learned on videos in the wild from unknown cameras, we collected videos from the YouTube8M dataset \cite{YouTube8M}. From the 3079 videos in YouTube8M that have the label ``quadcopter", human raters selected videos that contain significant amount of footage from a quadcopter. Naturally, the videos were taken with different unknown cameras, with varying fields of view and varying degrees of lens distortion. The YouTube8M IDs will be made public.

\subsection{Depth}
\paragraph{KITTI}
Table \ref{tab:kitti_eigen} summarizes the evaluation results on the KITTI Eigen partition of a model trained on KITTI. The metrics are the ones defined in Zhou et al. \cite{zhou2017unsupervised}. Only the best methods and the first three metrics are displayed in Table \ref{tab:kitti_eigen}, the rest are given in the Appendix.
As seen in the results, we obtain best state-of-the-art result. More importantly, we observed that learned intrinsics, rather than given ones consistently help performance, as seen too in later experiments.

\begin{table} [h!]
\small
  \centering
  {
  \begin{tabular}{|l|c|c|c|c|}
  \hline
  Method & M & Abs Rel & Sq Rel & RMSE  \\
  \hline 
  Zhou \cite{zhou2017unsupervised}&  & 0.208 & 1.768 & 6.856   \\
  Yang \cite{Yang2017unsupervised} &  &0.182 &1.481 &6.501  \\
  Mahjourian \cite{mahjourian2018unsupervised}&  & 0.163 & 1.240 & 6.220 \\ 
  LEGO \cite{yang2018lego} &\checkmark &0.162 &1.352 &6.276  \\
  GeoNet \cite{yin2018geonet}  &\checkmark  &0.155 &1.296 &5.857  \\
  DDVO \cite{wang2018learning} & &0.151 &1.257 &5.583  \\
  Godard \cite{godard2018digging}  &  &0.133 &1.158 &5.370 \\
  Struct2Depth \cite{casser2019struct2depth}  & \checkmark & 0.141  & 1.026  &5.291 \\
  Yang \cite{yang2018every} &  &0.137 &1.326 &6.232  \\
  Yang \cite{yang2018every}  &\checkmark  &0.131 &1.254 &6.117 \\

    \hline

  Ours: & & & & \\
  Given intrinsics &\checkmark & 0.129 & 0.982 & 5.23\\
  Learned intrinsics &\checkmark& \textbf{0.128} & \textbf{0.959} & \textbf{5.23} \\
  
  \hline
  \end{tabular}
  }
  \vspace{1mm}
  \caption{\small Evaluation of depth estimation of our method, with given and learned camera intrinsics, for models trained and evaluated on KITTI, compared to other monocular methods. The depth cutoff is always 80m. The ``M" column is checked for all models where object motion is taken into account.}
    \label{tab:kitti_eigen}
\end{table}

\begin{figure*}[h!]
\begin{center}
\includegraphics[width=0.23\linewidth]{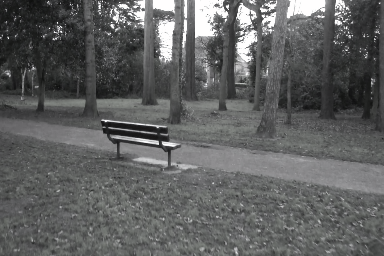}
\includegraphics[width=0.23\linewidth]{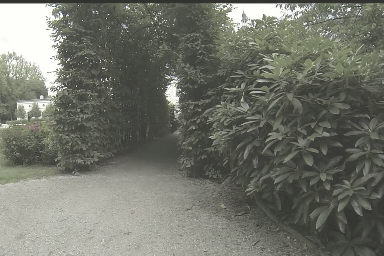}
\includegraphics[width=0.257\linewidth]{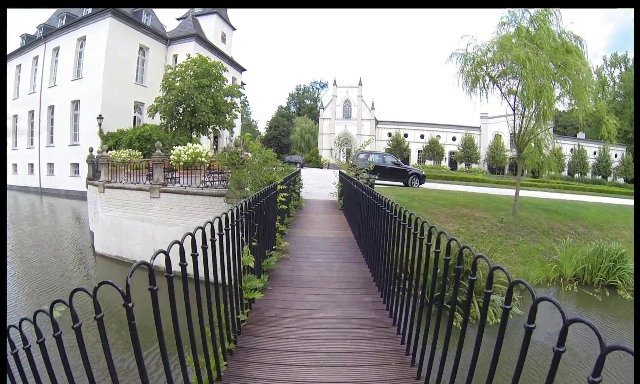}
\includegraphics[width=0.257\linewidth]{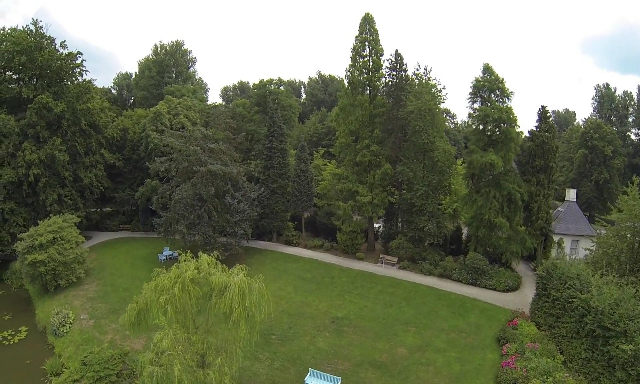}

\includegraphics[width=0.23\linewidth]{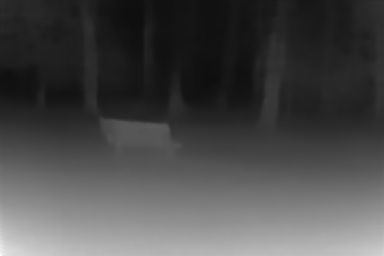}
\includegraphics[width=0.23\linewidth]{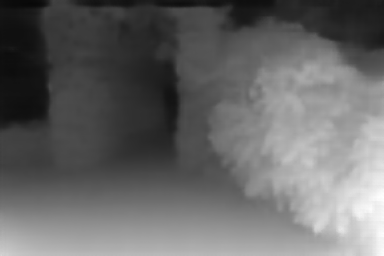}
\includegraphics[width=0.257\linewidth]{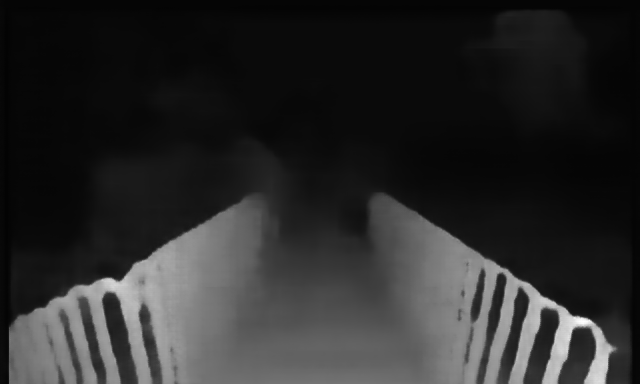}
\includegraphics[width=0.257\linewidth]{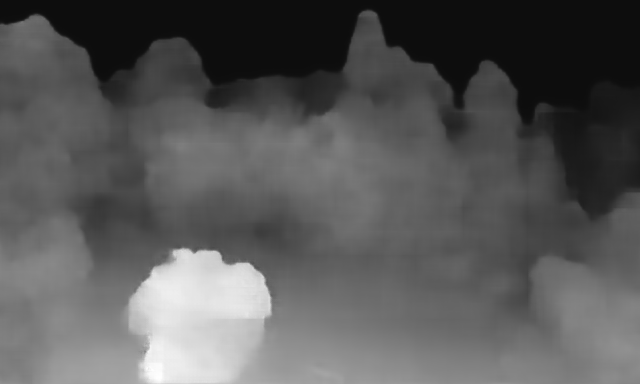}

\vspace{5mm}
\includegraphics[width=0.24\linewidth]{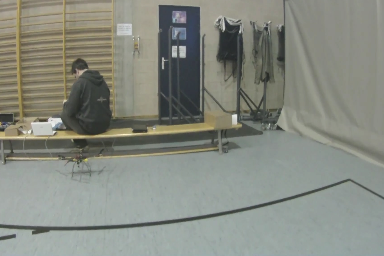}
\includegraphics[width=0.24\linewidth]{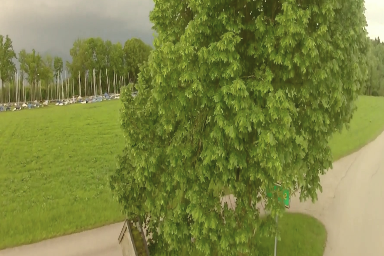}
\includegraphics[width=0.24\linewidth]{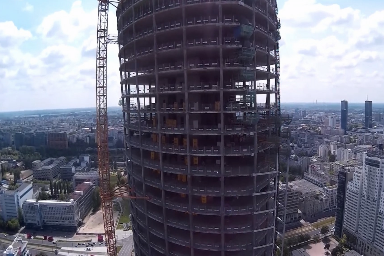}
\includegraphics[width=0.265\linewidth]{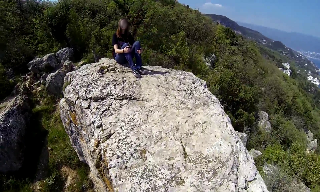}

\includegraphics[width=0.24\linewidth]{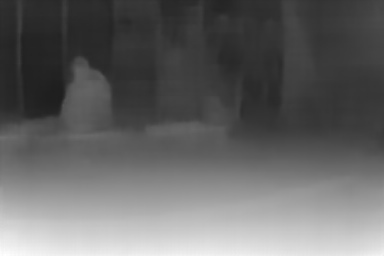}
\includegraphics[width=0.24\linewidth]{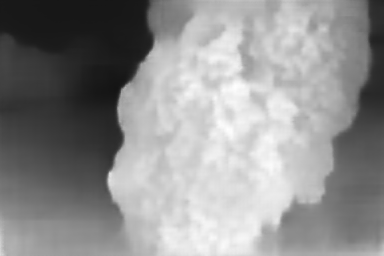}
\includegraphics[width=0.24\linewidth]{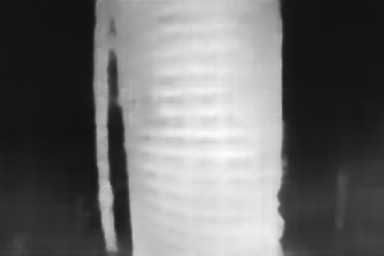}
\includegraphics[width=0.265\linewidth]{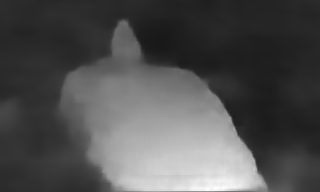}

\vspace{5mm}
\includegraphics[width=0.2453\linewidth]{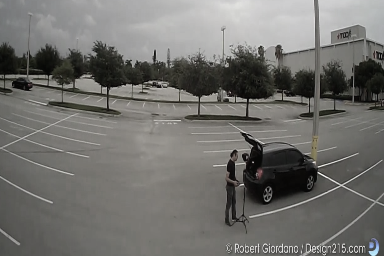}
\includegraphics[width=0.2453\linewidth]{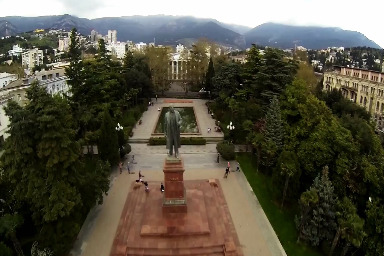}
\includegraphics[width=0.2453\linewidth]{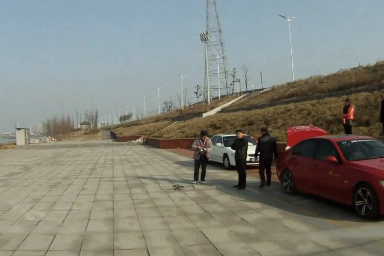}
\includegraphics[width=0.2453\linewidth]{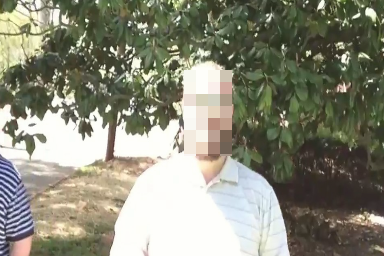}

\includegraphics[width=0.2453\linewidth]{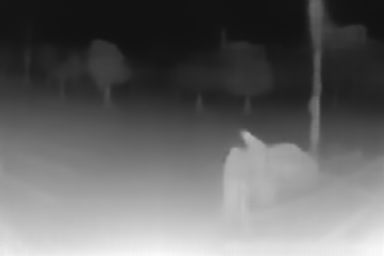}
\includegraphics[width=0.2453\linewidth]{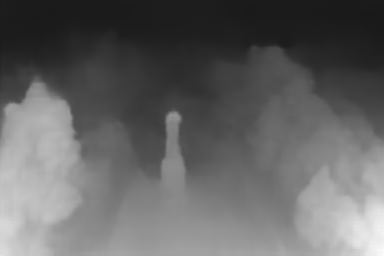}
\includegraphics[width=0.2453\linewidth]{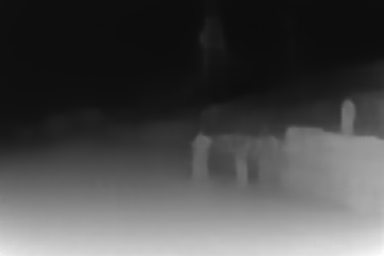}
\includegraphics[width=0.2453\linewidth]{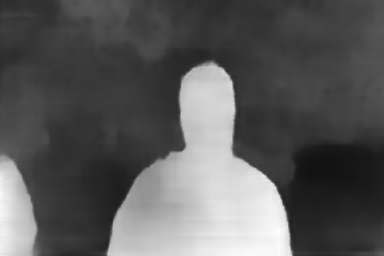}

\end{center}
  \caption{\small Images and learned disparity maps from the set collected from YouTube8M.}
\label{ytimages}
\end{figure*}

\paragraph{Cityscapes}
Table \ref{tab:city} summarizes the evaluation metrics of models trained and tested on Cityscapes. We follow the established protocol by previous work, using the disparity for evaluation~\cite{casser2019struct2depth,pilzer2018unsupervised}. Since this is a very challenging benchmark with many dynamic objects, very few approaches have evaluated on it. As seen in Table \ref{tab:city}, our approach outperforms previous ones and benefits from learned intrinsics.  

\begin{table} [h]
\small
  \centering
  {
  \begin{tabular}{|l|c|c|c|c|}
  \hline
  Method & M & Abs Rel & Sq Rel & RMSE  \\
  \hline 
  Pilzer \cite{Pilzer}& & 0.440& 5.713 & 5.443\\
  Struct2Depth \cite{casser2019struct2depth} & \checkmark  & 0.145 &1.736 &7.279\\ 
  \hline
  Ours: & & & & \\
  Given intrinsics &\checkmark &0.129 &1.35 & 6.96\\
  Learned intrsinsics &\checkmark &\bf 0.127 & \bf1.33 & \bf 6.96 \\
  \hline
\end{tabular}
  }
\vspace{1mm}
  \caption{\small Evaluation of depth estimation of models trained on Cityscapes on the cityscapes test set, with a depth cutoff of 80m, and comparison to prior art.}
    \label{tab:city}
\end{table}

\begin{figure}
\begin{center}
   \includegraphics[width=1\linewidth]{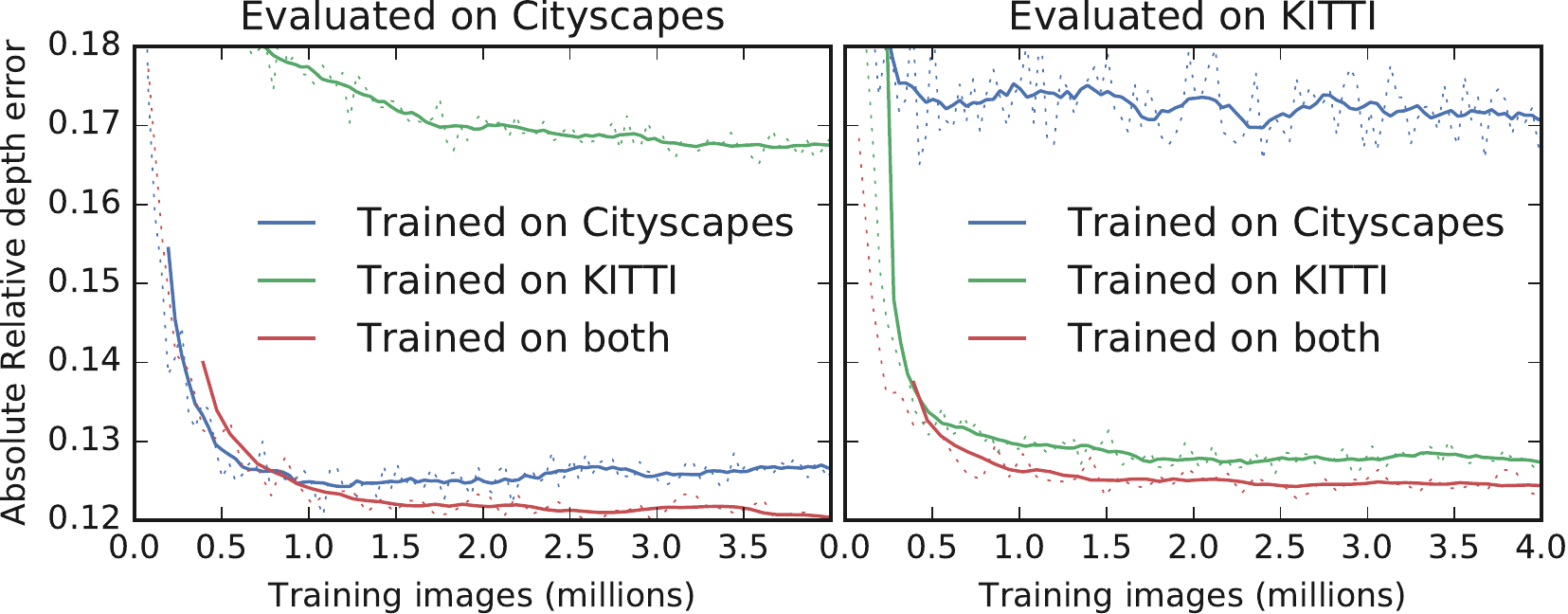}
\end{center}
   \caption{\small Depth prediction on Kitti and Cityscapes when training on each dataset and on both. Joining the two datasets improves the results on both. Learning intrinsics allows to similarly pool many datasets, even if they are from unknown cameras.}
\label{fig:depth_all}\end{figure}

\paragraph{Cityscapes + KITTI} Being able to learn depth without the need for intrinsics opens up the opportunity for pooling videos from any data sources together. Figure~\ref{fig:depth_all} shows the results of pooling Cityscapes and KITTI datasets and evaluating on either one, in this experiment the intrinsincs are assumed unknown and are learned. As seen jointly training improves the depth results even further than the best depth models on each dataset. This is a key result which demonstrates the impact of our proposed method to leverage  data sources of potentially unlimited size.

\paragraph{Cityscapes + KITTI: ablation experiments}
Table \ref{tab:ablation} summarizes the results of ablation experiments we ran in order to study the impact of each of the techniques described in this paper on the end results. In order to reduce the number of combinations of results, in all experiments the training set was Cityscapes and KITTI mixed together. Each model was evaluated on both Cityscapes and KITTI separately.

\begin{table} [ht]
\small
  \centering
  {
  \begin{tabular}{|l|c|c|}
  \hline  
    & \multicolumn{2}{c|}{Abs Rel depth}  \\
   Main Method & CS & KITTI \\
  \hline
  Our algorithm & 0.121& \textbf{0.124}\\ 
  \hline
  Boxes instead of masks & \textbf{0.120} & 0.125 \\
  w/o occlusion-aware loss & 0.127 & 0.126 \\
  w/o object motion & 0.172 &  0.130 \\
  w/o randomized layer normalization & 0.124 & 0.127\\
  \hline
  \end{tabular}
  } 
\vspace{1mm}
  \caption{\small Ablation experiments on depth estimation. In all experiments the training set was Cityscapes (CS) and KITTI combined, and we tested the model on Cityscapes (CS) and KITTI (Eigen partition) separately. Each row represents an experiment where one change was made compared to the main method, as described in the ``Experiment" row. Smaller numbers are better.}
    \label{tab:ablation}
\end{table}

Using a union of bounding boxes as a ``possibly-mobile" mask, as depicted in Fig.~\ref{fig:boxify} is found to be as good as using segmentation masks, which makes our technique more broadly applicable. Object motion estimation is shown to play a crucial role, especially on Cityscapes, which is characterized by more complex scenes with more pedestrians and cars. Randomized layer noralizarion is shown to be superior to standard batch normalization, and at last -- accounting for occlusions significantly improves the quality of depth estimation, especially on Cityscapes, which has richer scenes with more occlusions. Figure~\ref{occlusion_ablation} visualizes the type of artifacts that occlusion-aware losses reduce.

\begin{figure}[ht]
\begin{center}
   \includegraphics[width=0.8\linewidth]{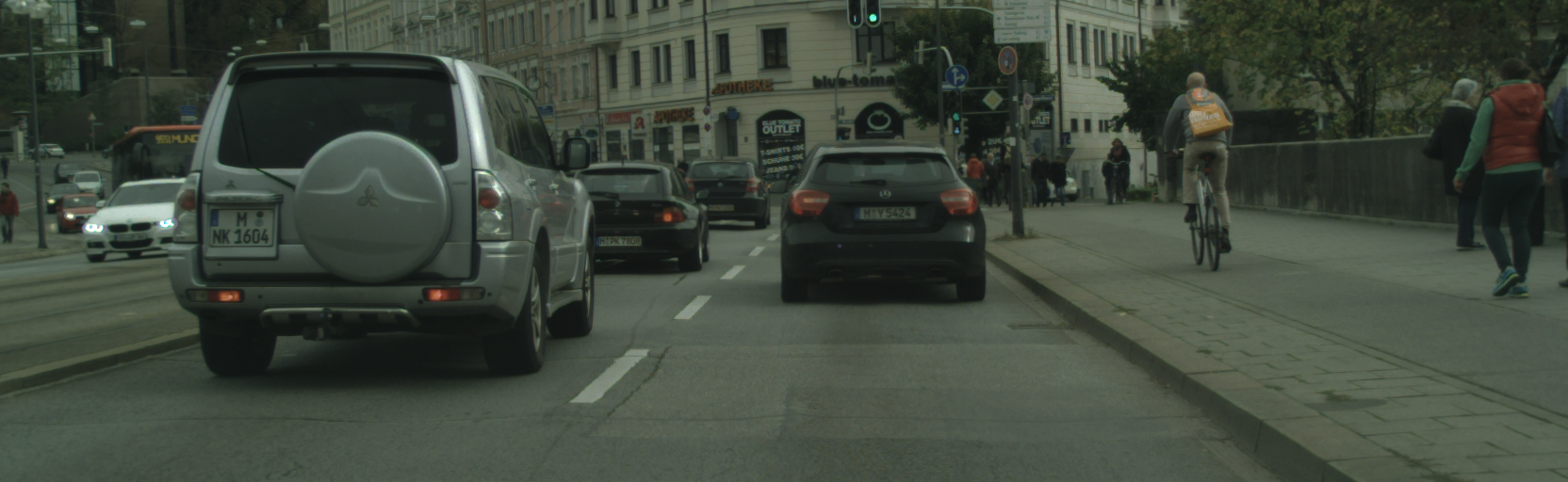}
\includegraphics[width=0.8\linewidth]{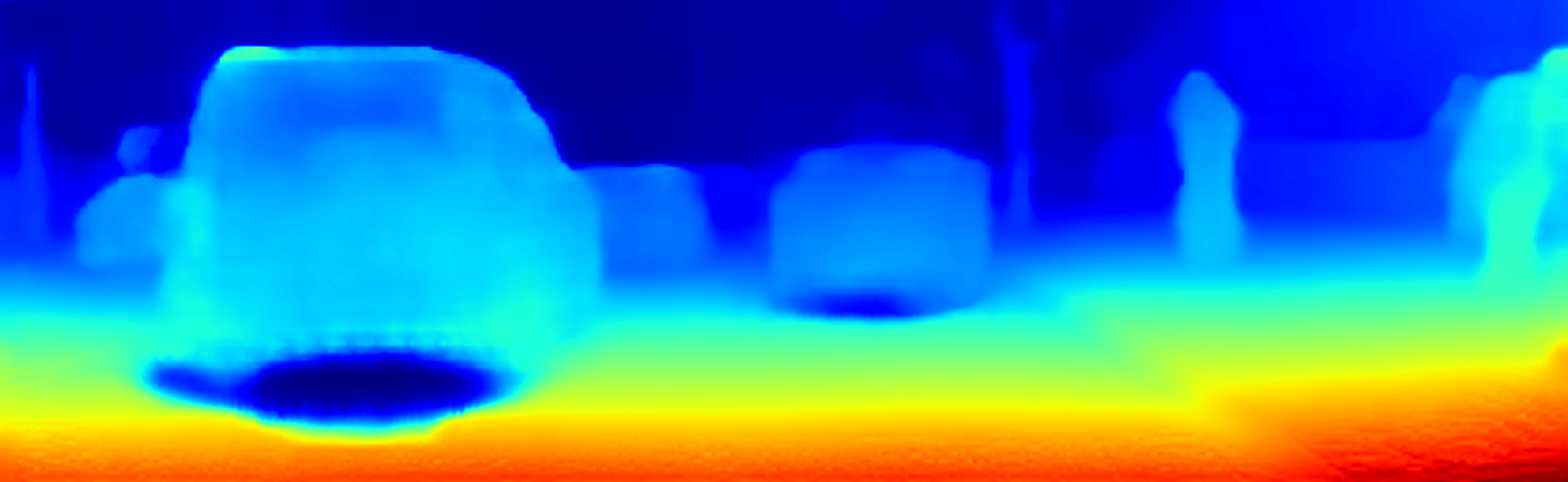}
\includegraphics[width=0.8\linewidth]{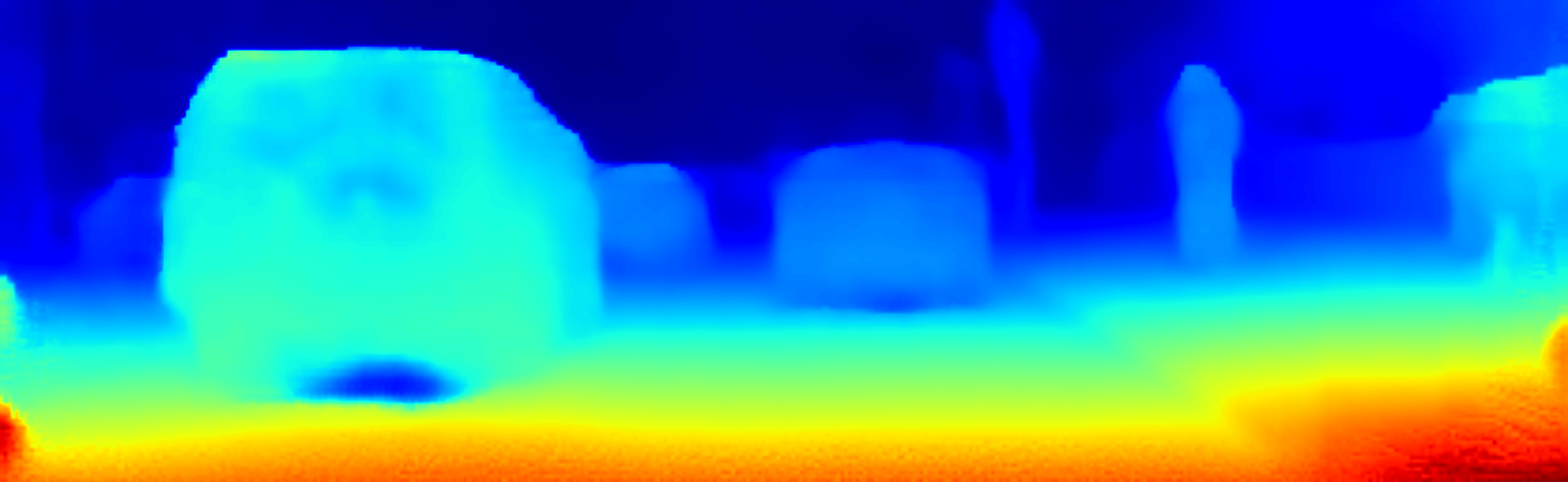}
      
\end{center}
   \caption{\small Illustration of the effect of occlusion aware losses. The center and bottom images are inferred disparity maps obtained from the image at the top. In the center image, the model was trained without occlusion-aware losses. At areas that become disoccluded, under cars, occlusion aware loss is shown to reduce artifacts. The top image belongs to the Cityscapes test set and is one of images that whose depth prediction was hurt the most upon removing occlusion aware losses.}
\label{vicondepth}
\end{figure}

\paragraph{EuRoC MAV Dataset}
We further use the EuRoC MAV Dataset to evaluate depth. 
We selected also a very challenging out-of-sample evaluation protocol in which we trained on the ``Machine room" sequences and tested on the ``Vicon Room 2 01, which has 3D groundtruth. Table~\ref{tab:euroc} reports the results. The Appendix details how depth ground truth was generated from the provided Vicon 3D scans.

\begin{figure}[ht]
\begin{center}\label{ytdepth}
   \includegraphics[width=1.0\linewidth]{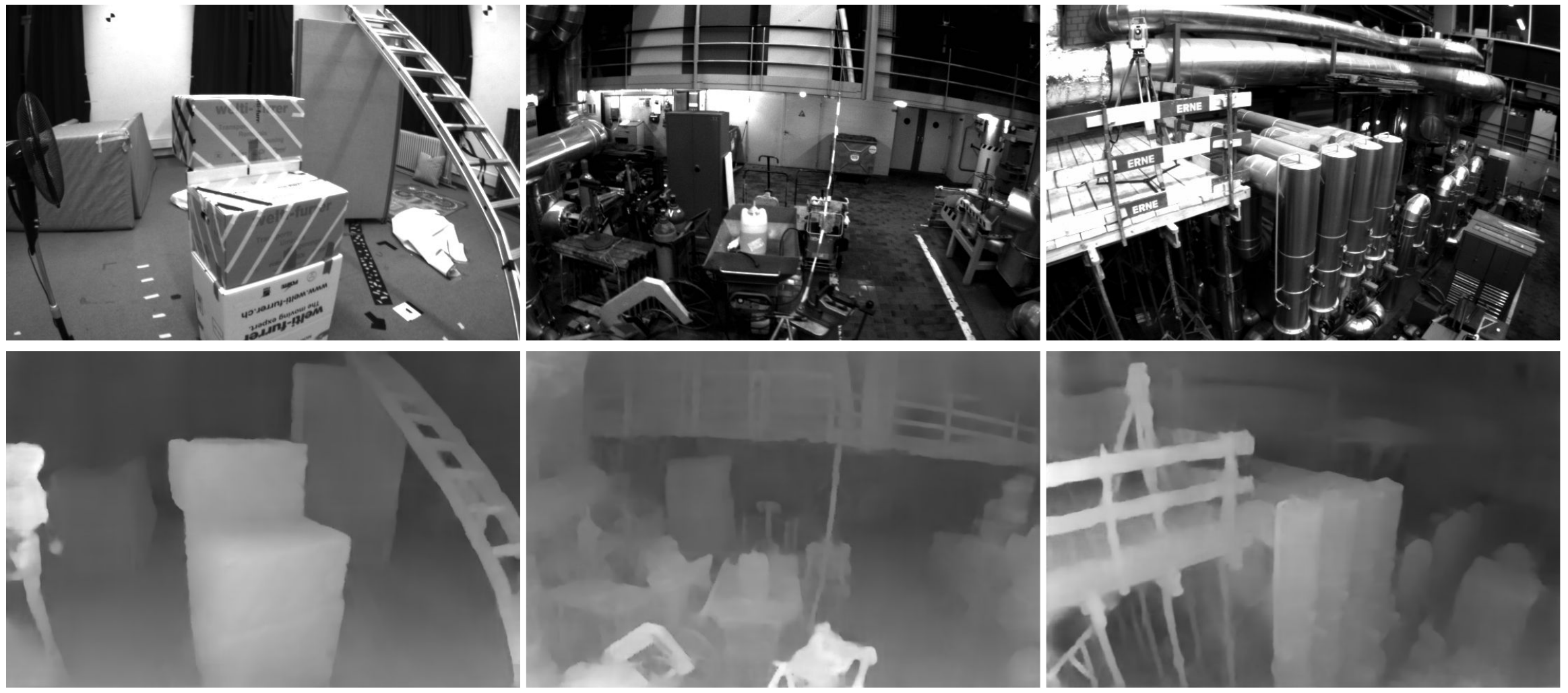}
\end{center}
  \caption{\small Frames from EuRoC~\cite{Burri2016Euroc} and the corresponding learned disparity maps. The camera intrinsics are learned.}
  \label{fig:euroc_im}
\end{figure} 

\begin{table} [h]
\small
  \centering
  {
  \begin{tabular}{|c|c|c|c||c|c|c|}
  \hline
   Abs R  & Sq R & RMS & logRMS &$a_1$ &$a_2$ &$a_3$ \\
  \hline
   0.332    &0.389    &0.971    &0.396    &0.420    &0.743  &  0.913\\

  \hline
  
\end{tabular}
  }
  \vspace{1mm}
  \caption{\small Evaluation of depth estimation for EuRoC dataset, no prior art is available for this dataset. $a_i = \delta <1.25^i$.}
    \label{tab:euroc}
\end{table}

Figure~\ref{fig:euroc_im} visualizes some of the learned depth. The EuRoC results were obtained from a network trained on the entire EuRoC dataset. Intrinsics are learned, and as seen later are very close to the real ones. No other information is used as input.  
 
\subsection{YouTube Videos}
\begin{figure} [ht]
\begin{center}\label{ytdepth}
    \includegraphics[width=1.0\linewidth]{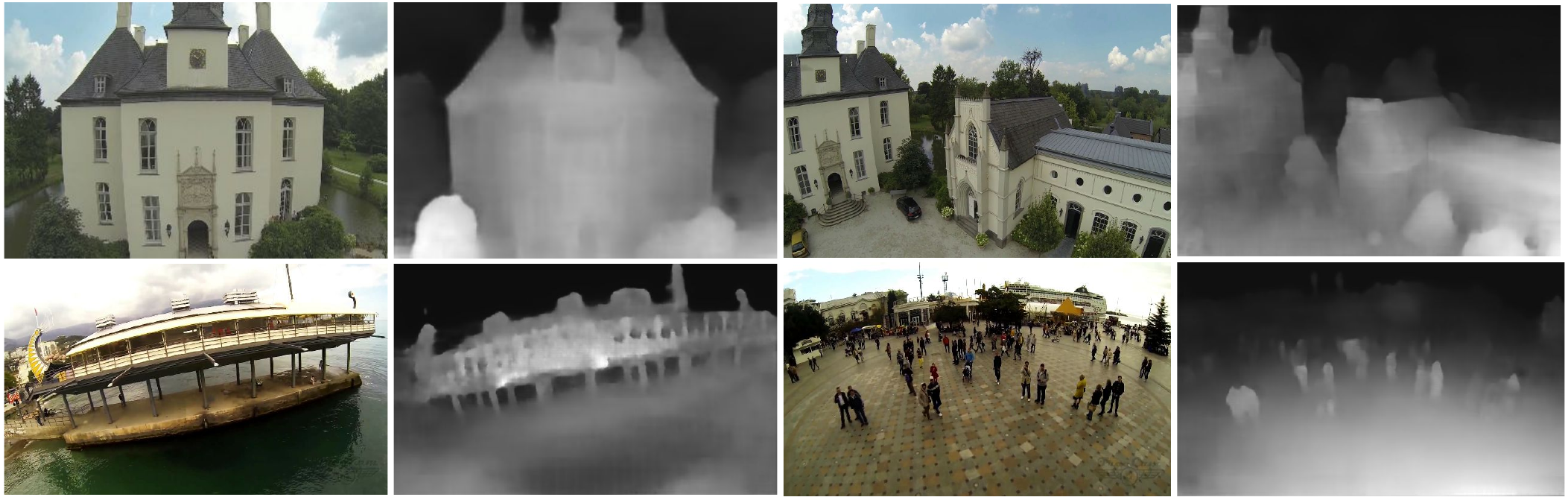}
\end{center}
  \caption{\small Frames from YouTube and learned disparity maps. The camera intrinsics are learned. 
  }
  \label{fig:yt}
\end{figure}

To demonstrate that depth can be learned on collections of videos in the wild for which the camera parameters are not known, and differ across videos, traained our model on the YouTube8M videos described in Sec.~\ref{sec:datasets}. Figure~\ref{fig:yt} visualizes the results. We note that this dataset is very challenging as it features objects at large ranges of depth. 

\subsection{Camera intrinsics evaluation}
In order to evaluate our method with respect to learning camera intrinsics, two separate questions can be asked. First, how good is the supervision signal that cross-frame consistency provides for the camera intrinsics. Second is how good a deep network is in learning them and generalizing to test data.

\paragraph{Quality of the supervision signal.}
To estimate the supervision signal for the intrinsics, we represented each intrinsic parameter as a separate learned variable. The model was trained on the EuRoc dataset, since it was captured with the same camera throughout. Each of the 11 subsets of EuRoC was trained in a separate experiment, until the convergence of the intrinsics variables, yielding 11 independent results. The resulting sample mean and standard deviation for each intrinsic parameter are summarized in Table \ref{tab:EuRocIntrinsics}. All parameters agree with groundtruth within a few pixels. Since the groundtruth values were not accompanied by tolerances, it is hard to tell if the differences are within tolerance or not.
\begin{table} [h]
\small
  \centering
  {
  \begin{tabular}{|l|c|c|}
  \hline
  Quantity & Learned & GT \\
  \hline
  Horizontal focal length ($f_x$)& $253.7 \pm 1.1$ & $250.2$\\
  Vertical focal length ($f_y$) & $265.4 \pm 1.3$ & $261.3$\\
  Horizontal center ($x_0$) & 189.0 $\pm$ 0.9  & 187.2 \\
  Vertical center ($y_0$) & 132.2 $\pm$ 1.1  & 132.8 \\
  Quadratic radial distortion& $-0.267 \pm 0.003$ & $-0.283\phantom0$ \\
  Quartic radial distortion& $\phantom-0.064 \pm 0.002$ & $0.074$ \\
  \hline
  \end{tabular}
  }
\vspace{1mm}
  \caption{\small Camera intrinsics learned on the EuRoC datasets. Learning of depth, egomotion and intrinsics was done separately on each of the 11 datasets, using monocular images (``cam0") only. Constancy of the intrinsics throughout each dataset separately was imposed, and statistics (mean and standard deviation) for each intrinsic parameter were collected across the results. The groundtruth (GT) was adjusted to cropping and resizing used in our training, to a size of 256$\times$384. 
}
    \label{tab:EuRocIntrinsics}
\end{table}

\paragraph{Learning and generalization} Prior work \cite{workman2015deepfocal,FocalLens,Bogdan_2018} has shown that deep networks can learn and generalize camera intrinsics in a strongly supervised setting. In our setting, the camera intrinsics and motion are predicted by the same network and are thus correlated. In other words, the losses imposed on the motion / intrinsics network only impose correctness of the intrinsics within the limits of Eq.~\ref{fcond}. This is the downside of having the same network predict both quantities, the advantage of which is reduced computational cost. 

\begin{figure}[ht]
\begin{center}
   \includegraphics[width=0.9\linewidth]{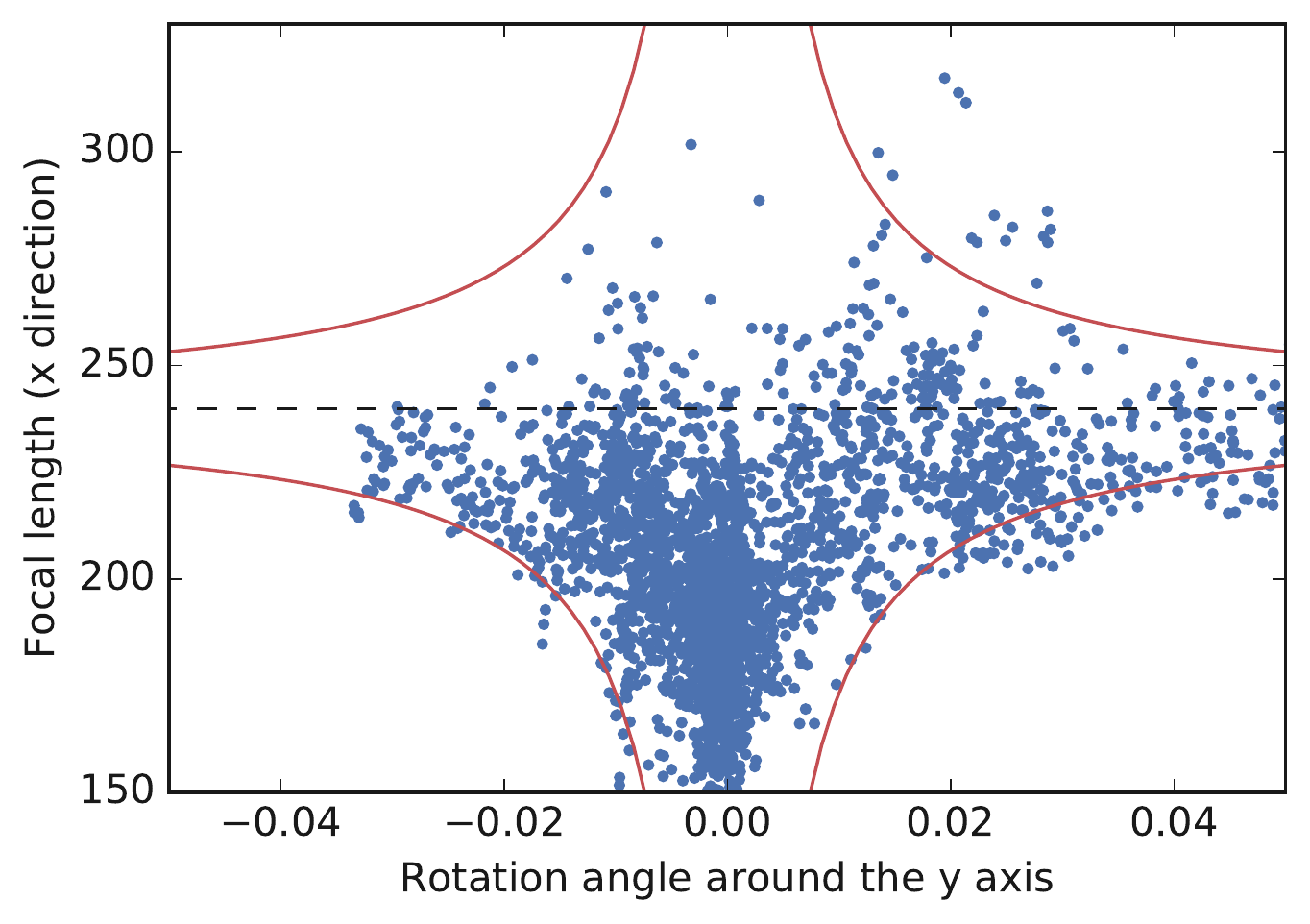}
\end{center}
   \caption{\small Predicted $f_x$ as function of the predicted $r_y$ for all images in the KITTI odometry sequences 09 and 10. The dashed line is groundtruth; red curves show tolerance limits imposed by Eq.~\ref{fcond}.}
\label{fxfig}\end{figure}

 We evaluated our model's predictions of the intrinsic parameters on the KITTI odometry series. The model was trained on the union of the Cityscapes and KITTI training sets. Figure \ref{fxfig} shows the scatter plot of the predicted $f_x$ as function of the predicted $r_y$. The predictions fall within the limits imposed by Eq.~\ref{fcond}. 
 While Table~\ref{tab:EuRocIntrinsics} indicates a high-quality supervision signal for the intrinsics, Fig.~\ref{fxfig} shows that in the setting where the intrinsics and motion are predicted by the same network, the latter learns them ``in aggregate", only to the extent needed for the depth-prediciton network to learn depth.

\subsection{Odometry}
We evaluated our egomotion prediction on the KITTI sequences 09 and 10. The common 5-point Absolute Trajectory Error (ATE) metric \cite{zhou2017unsupervised, casser2019struct2depth, yin2018geonet, godard2018digging} measures local agreement between the the estimated trajectories and the respective groundtruth. However assessing the usefulness for a method for localization requires evaluating its accuracy in predicting location. A common metric for that is average relative translational drift $t_{rel}$ \cite{ShamwellOdometry, zhan2018unsupervised} - the distance between the predicted location and the groundtruth location divided by distance traveled and averaged over the trajectory. Table \ref{tab:odoboth} summarizes both metrics, demonstrating the improvements our method achieves on both. 

\begin{figure}[ht]
\begin{center}
      \includegraphics[width=0.9\linewidth]{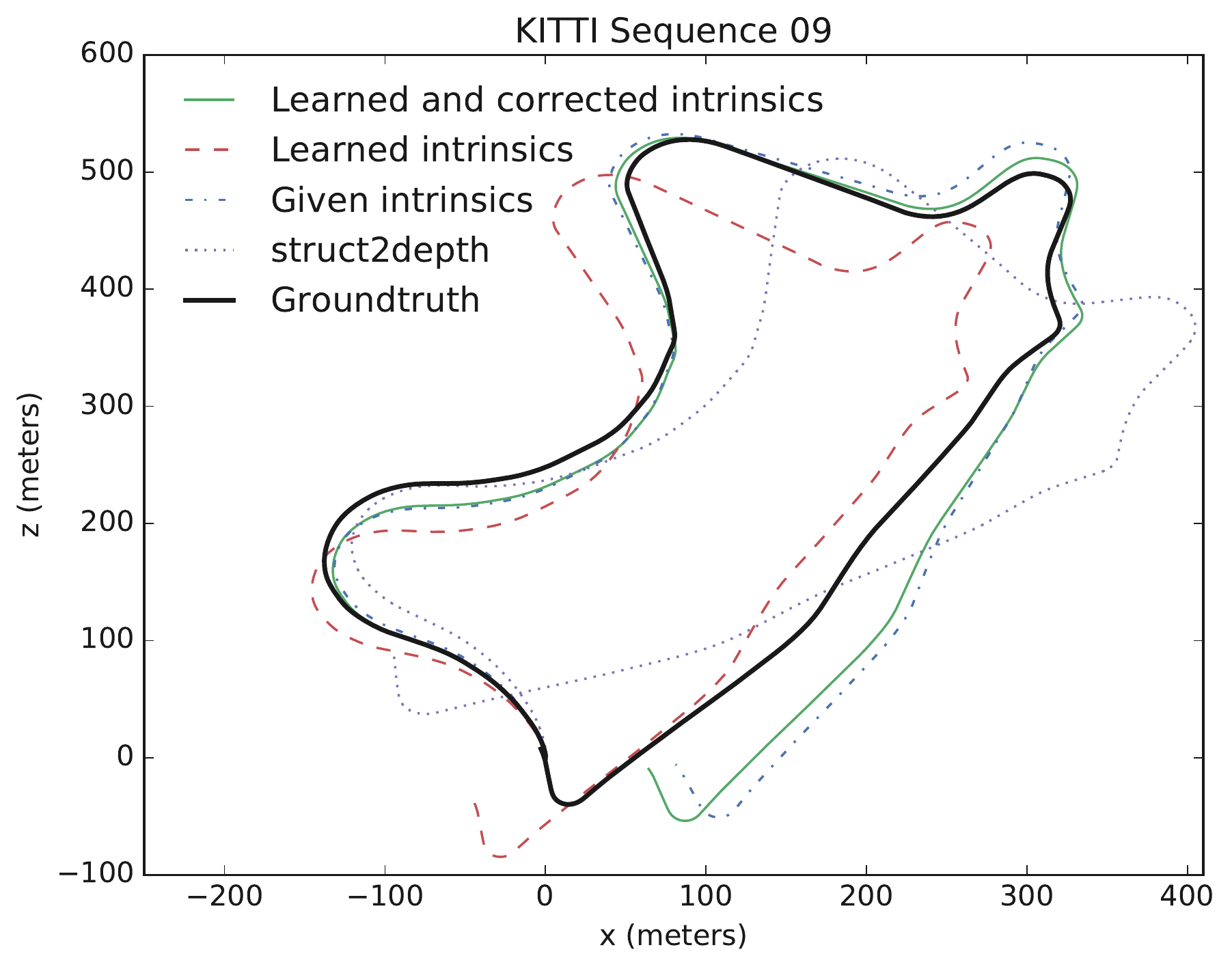}
\end{center}
   \caption{\small Predicted location on the KITTI odometry sequence 09, generated by models trained on KITTI, with given intrinsics and with learned intrinsics (with and without inference time correction), compared to groundtruth and to struct2depth results.}
\label{odofig}
\end{figure}

When evaluating for odometry, the most naive way is to calculate the inference of egomotion for every pair of adjacent frame. That leads to the red ``learned intrinsics" curve in Fig.~\ref{odofig}. However it is also possible to make an inference-time correction if we know the intrinsics of the camera at inference time. In that case, one can leverage the fact that for small errors in the rotation angles and focal lengths, $r_x f_y$ and $r_y f_x$ are approximately constant. Therefore if the network predicted $r'_y$ and $f'_x$ for a given pair of images, and we know the true focal length $f_x$, we can correct our estimate of $r_y$ to $r'_y f'_x / f_x$. This is the procedure invoked in generating the ``Learned and corrected intrinsics" curve in Fig.~\ref{odofig}, and the respective entry in Table~\ref{tab:odoboth}. The trajectories and metrics obtained with learned itrinsics with inference time correction and with given intrinsics are similar. Both notably improve prior art, which is especially prominent for localization, as the $t_{rel}$ metric indicates.

\begin{table} [h!]
  \centering
  \small
  {
  \begin{tabular}{|l||c|c||c|c|}
  \hline
    &\multicolumn{2}{c||}{\underline{Seq.~09}} & \multicolumn{2}{c|}{\underline{Seq.~10}} \\
   Metric & ATE & $t_{rel}$ & ATE & $t_{rel}$ \\
  \hline 
  Zhou \cite{zhou2017unsupervised} & 0.021 & 17.84\%& 0.020 & 37.91\%\\
  GeoNet \cite{yin2018geonet} &  0.012 & / & 0.012 & /  \\
  
  Zhan \cite{zhan2018unsupervised} & / & 11.92\%& /  &12.45\%\\
  Mahjourian \cite{mahjourian2018unsupervised} & 0.013 & / & 0.012 & / \\
  Struct2depth \cite{casser2019struct2depth} & 0.011 & 10.2\%&0.011& 28.9\%\\
  \hline
  Ours, with intrinsics: &&&&\\ 
  Learned  & 0.012  &7.5\%& 0.010 & 13.2\%\\
  Learned \& corrected & 0.010  &\textbf{2.7\%}& \textbf{0.007} & 6.8\%\\
  Given & \textbf{0.009} & 3.1\%& 0.008 & \textbf{5.4\%}\\
  \hline
  \end{tabular}
  }
  \vspace{1mm}
  \caption{\small Absolute Trajectory Error (ATE) \cite{zhou2017unsupervised} and average relative translational drift ($t_{rel}$) \cite{ShamwellOdometry} on the 09 and 10 KITTI odometry sequences. Our method with both learned and given intrinsics is compared to prior work.
  }
    \label{tab:odoboth}
\end{table}
\section{Conclusions}
This work addresses important challenges for unsupervised learning of depth and visual odometry through geometric handling of occlusions, a simple way of accounting for object motion, and a novel form of regularization. Most importantly, it takes a major step towards leveraging the vast amounts existing unlabeled videos for learning depth estimation: Through unsupervised learning camera intrinsic parameters, including lens distortion, it enables for, the first time, learning depth from raw videos captured by unknown cameras.

\section{Acknowledgements}
We Thank Roland Siegwart for the permission to use EuRoC dataset, Sergey Ioffe for fruitful discussions, and Kurt Konolige for consultation and critical reading of the manuscript.

{\small
\bibliographystyle{ieee}
\bibliography{egbib}
}

\setcounter{section}{0}
\renewcommand{\thesection}{\Alph{section}}

\setcounter{equation}{0}
\renewcommand{\theequation}{A\arabic{equation}}

\setcounter{table}{0}
\renewcommand{\thetable}{A\arabic{table}}

\setcounter{figure}{0}
\renewcommand{\thefigure}{A\arabic{figure}}

\clearpage
\section{Appendix}

\subsection{Accuracy of camera intrinsics - derivation}
In Section 4.1 of the main paper we showed that if $K$ is predicted incorrectly ($\tilde K$), there is an incorrect value of $t$, $\tilde t$, such that $\tilde K \tilde t$ is still correct. The situation is different with rotations. Let $K$ and $R$ be the correct intrinsic matrix, and $\tilde K$ and $\tilde R$ be inaccurate ones, predicted by the network. If
\begin{equation}\label{rtilde}
KRK^{-1}=\tilde K\tilde R\tilde K^{-1},
\end{equation}
according to Eq.~(1) the training loss will not be affected. Proving and analyzing this statement is the subject of this section.

\subsection{If $\bf R\neq I$, $\bf \tilde K = K$: Proof}
Isolating $\tilde R$ from Eq.~(\ref{rtilde}) and asserting $\tilde R \tilde R^T=I$, we obtain
\begin{equation}\label{ARRA}
    AR = RA, \quad \mathrm{where}\quad  A = K^{-1}\tilde K \tilde K^T K^{T^{-1}}
\end{equation}
If $R$ is the identity matrix $I$, that is, there is no rotation, Eq.~\ref{ARRA} holds trivially, which means that no supervision dsignal is passed to $\tilde K$. We henceforth assume $R\neq I$.

The determinant of $A$ is 1, and from calculating the characteristic polynomial of $A$, we can show that $A$ always has 1 as an eigenvalue (this can be done using a symbolic math package). Therefore either $A$ has 3 distinct eigenvalues, or all its eigenvalues are equal one, which means it is the identity matrix. The former option leads to a contradiction, because since A is symmetric and real, its eigenvectors can be chosen all real. Since $AR=RA$, $R$ must share the same eigenvectors. However SO(3) matrices that are not the identity matrix and nor a rotation in 180 degrees nor a reflection\footnote{Rotations by 180 degrees or reflections are not relevant in the context of transforming a frame to an adjacent one.}, have necessarily complex eigenvectors. Therefore $A$ must be the identity matrix, which leads to $\tilde K \tilde K^T = KK^T$. Substituting $K$ from Eq.~(2), it follows that $K = \tilde K$.

\subsection{Tolerances}
Mathematically, $R=I$ provides no supervision signal for $K$, whereas $R\neq I$ provides a complete supervision siglal for it. Practically, it is clear that the closer $R$ is to $I$, the weaker the supervision on $K$ is. Here we would like to quantify this relation.

For simplicity, suppose we have only rotation around $y$ and no translation. Eq.~(1) then reduces to
\begin{equation}\label{deltapx}
\Delta p_x = r_y f_x + \frac{r_y(p_x - x_0)^2}{f_x},
\end{equation}
where $\Delta p_x = p'_x-p_x$. $\Delta p_x$ is where the supervision signal comes from.

Let $\tilde f_x$ be an erroneous prediction of $f_x$ by a deep network. Right in front of the pinhole, which is approximately at the center of the image\footnote{The assumption that the pinhole is at the center of the image is only used for simplicity of the derivation here. It is not assumed during training, and is not true, for example, in the EuRoC set.}, $p_x=x_0$ and the error in $\Delta p_x$ can be fully compensated for by predicting 
\begin{equation}\label{rprime}
\tilde r_y = r_y f_x / \tilde f_x.
\end{equation}
Replacing $f_x$ and $r_y$ by $\tilde f_x$ and $\tilde r_y$ in Eq.~\ref{deltapx}, and using Eq.~\ref{rprime} to eliminate $\tilde r_x$, we obtain 
\begin{equation}\label{deltapxp}
\Delta \tilde p_x = r_y f_x + \frac{r_y f_x(p_x - x_0)^2}{\tilde f_x^2}
\end{equation}
and 
\begin{equation}\label{deltapxd}
\Delta \tilde p_x - \Delta p_x = (p_x - x_0)^2 r_y f_x\left(\frac{1}{\tilde f_x^2}-\frac{1}{f_x^2}\right)
\end{equation}

Let $\delta f_x$ and $\delta r_y$ be the errors in estimating $\tilde f_x - f_x$ and $\tilde r_y - r_y$ respectively. To first order in $\delta f_x$ and $\delta r_y$, Eq.~\ref{deltapxd} becomes
\begin{equation}\label{ddeltapx1}
\Delta \tilde p_x - \Delta p_x = \frac{2(p_x - x_0)^2 r_y}{ f^2_x}(f_x - \tilde f_x)
\end{equation}

\begin{table*}[t]
\small
  \centering
  {
  \begin{tabular}{|l|c|c|c|c|c||c|c|c|}
  \hline
  Method & M & Abs Rel & Sq Rel & RMSE & RMSE log & $\delta < 1.25$ & $\delta < 1.25^2$ & $\delta < 1.25^3$ \\
  \hline 
  Zhou \cite{zhou2017unsupervised}&  & 0.208 & 1.768 & 6.856 & 0.283& 0.678 & 0.885& 0.957 \\
  Yang \cite{Yang2017unsupervised} &  &0.182 &1.481 &6.501 &0.267 &0.725 &0.906 &0.963 \\
  Mahjourian \cite{mahjourian2018unsupervised}&  & 0.163 & 1.240 & 6.220 & 0.250 &0.762 &0.916 &0.968\\ 
  LEGO \cite{yang2018lego} &\checkmark &0.162 &1.352 &6.276 & 0.252 &0.783& 0.921 &0.969 \\
  GeoNet \cite{yin2018geonet}  &\checkmark  &0.155 &1.296 &5.857& 0.233 &0.793 &0.931 &0.973  \\
  DDVO \cite{wang2018learning} & &0.151 &1.257 &5.583 & 0.228 &0.810 &0.936 &0.974 \\
  Godard \cite{godard2018digging}  &  &0.133 &1.158 &5.370 & \textbf{0.208} &0.841 & \textbf{0.949} &0.978\\
  Struct2Depth \cite{casser2019struct2depth}  & \checkmark & 0.141  & 1.026  &5.291 &0.2153 &0.8160 &0.9452 &\textbf{0.9791} \\
  Yang \cite{yang2018every} &  &0.137 &1.326 &6.232&0.224 &0.806 &0.927 &0.973  \\
  Yang \cite{yang2018every}  &\checkmark  &0.131 &1.254 &6.117 &0.220 &0.826 &0.931 &0.973  \\

    \hline

  Ours: & & & & & & &  &\\
  Given intrinsics &\checkmark & 0.129 & 0.982 & 5.23 & 0.213 & 0.840 & 0.945 & 0.976\\
  Learned intrinsics &\checkmark& \textbf{0.128} & \textbf{0.959} & \textbf{5.23} & {0.212} &  \textbf{0.845} & 0.947 & 0.976\\
  
  \hline
  \end{tabular}
  }
  \vspace{1mm}
  \caption{\small Evaluation of depth estimation of our method, with given and learned camera intrinsics, for models trained and evaluated on KITTI, compared to other monocular methods. The depth cutoff is always 80m. The ``M" column is checked for all models where object motion is taken into account. This extends Table \ref{tab:kitti_eigen} in the main paper. }
    \label{a1}
\end{table*}

\begin{table*} [h]
\small
  \centering
  {
 \begin{tabular}{|l|c|c|c|c|c||c|c|c|}
  \hline
  Method & M & Abs Rel & Sq Rel & RMSE & RMSE log & $\delta < 1.25$ & $\delta < 1.25^2$ & $\delta < 1.25^3$ \\
  \hline 
  Pilzer \cite{Pilzer}& & 0.440& 5.71 & 5.44 & 0.398 & 0.730 & 0.887 & 0.944\\
  Struct2Depth  \cite{casser2019struct2depth} & \checkmark  & 0.145 &1.74 &7.28 & 0.205 &0.813 & 0.942 & 0.978 \\ 
  \hline
  Ours: & & & & & & & & \\
  Given intrinsics &\checkmark &0.129 &1.35 & 6.96 &0.198 &0.827 & 0.945& 0.980\\
  Learned intrsinsics &\checkmark &\bf 0.127 & \bf1.33 & \bf 6.96 & \textbf{0.195} &\textbf{0.830}  &\textbf{0.947}& \textbf{0.981} \\
  \hline
\end{tabular}
  }
\vspace{1mm}
  \caption{\small Evaluation of depth estimation of models trained on Cityscapes on the cityscapes test set using the procedure and code in Ref.~\cite{casser2019struct2depth}, with a depth cutoff of 80m, and comparison to prior art. This table extends Table 2 from the main paper.}
  \label{a2}
\end{table*}

\begin{table*} [h!] 
\small
  \centering
  {
 \begin{tabular}{|l|c|c|c|c|c||c|c|c|}
  \hline
  Trained on & Evaluated on & Abs Rel & Sq Rel & RMSE & RMSE log & $\delta < 1.25$ & $\delta < 1.25^2$ & $\delta < 1.25^3$ \\
  \hline 
  Cityscapes &Cityscapes &0.127 & 1.33 &  6.96 & {0.195} &{0.830}  &{0.947}& {0.981}\\
    Cityscapes &KITTI & 0.172& 1.37& 6.21& 0.250 &0.754 & 0.921&0.967\\
    \hline
    KITTI &Cityscapes &0.167 &2.31 &9.99 & 0.272 &0.747 &0.894 &0.957\\
      KITTI &KITTI & {0.128} & {0.959} & {5.23} & {0.212} &  {0.845} & 0.947 & 0.976\\
      \hline
    Cityscapes +  KITTI &Cityscapes & \bf0.121 &\bf1.31 &\bf6.92 &\bf0.189 &\bf0.846 &\bf0.953 &\bf0.983\\
        Cityscapes +  KITTI &KITTI &\bf0.124 &\bf0.930 &\bf5.12 &\bf0.206 &\bf0.851 &\bf0.950 & \bf0.978\\
  \hline
\end{tabular}
  }
\vspace{1mm}
  \caption{\small Evaluation of depth estimation of models trained on Cityscapes and kitti together, on the cityscapes and KITTI test sets separately. The depth cuttof is of 80m. This table extends Figure \ref{fig:depth_all} in the main paper.}
     \label{f4}
\end{table*}

The worst case of Eq.~\ref{ddeltapx1} is where $p_x = w/2$:
\begin{equation}\label{ddeltapx2}
\Delta \tilde p_x - \Delta p_x < \frac{w^2 r_y}{2 f^2_x}(f_x - \tilde f_x)
\end{equation}

As long as the absolute value of the right hand side of Eq.~\ref{ddeltapx2} is $\ll 1$, the worst case error in pixel coordinates is much less than one pixel. This gives the condition
\begin{equation}\label{fxcond}
|\delta f_x| \ll \frac{2f_x^2}{w^2 r_y}
\end{equation}
A similar relation can be derived for $\delta f_y$. This is how Eq.~(3) was established.

\subsection{Full tables of metrics for depth estimation}

The numbers in  Table \ref{tab:kitti_eigen} and \ref{tab:city}, as well as in Fig.~\ref{fig:depth_all}, are given for only part of the metrics commonly published for depth estimation. In this section we give the rest of the metrics, for completeness. Tables \ref{a1}, \ref{a2} and \ref{f4} provide the full set of numbers for the former ones, respectively.

\subsection{Generating depth groundtruth for the EuRoC dataset}
In the EuRoC dataset, the Vicon Room 2 series has pointclouds that were obtained from merging depth sensor readings. In addition, there is groundtruth for the position and orientation of the camera at given timestamps, as well as the intrinsics. For every frame, we reprojected the point clouds onto the camera using the intrinsics and extrinsics. To address occlusions, each point was given some finite angular width. If two 3D points were projected onto close enough locations on the image plane, and their depth ratio was greater than a certain threshold, only the one closer to the camera was kept. Finally, the rendered depth maps were made more uniform by introducing a uniform grid in projective  space and keeping at most one point in a each cell. The code performing the above transformation will be released. An example of the result is shown in Fig.~\ref{vicondepth}.

\subsection{Intrinsics transformation on the EuRoC dataset}
The intrinsics of cam0 in the EuRoC set are $(752, 480)$ for the width and height, $458.654$, $457.296$ for the focal lengths in the $x$ and $y$ direction respectively, and $367.215$, $248.375$ for $x_0$ and $y_0$ respectively. 
The radial distortion coefficients are -0.28340811 and 0.07395907, and the higher-order coefficients are small. In our experiments, we first center-cropped the images to (704, 448). 
This does not change the focal lengths nor the distortion coefficients, and changes $x_0$ and $y_0$ to 343.215, 232.375
respectively. Next, we resized the images to (384, 256), which multiplies all $x$-related parameters by $284/704$, and all $y$-related parameters by $256/448$. The results are in the last column of Table 5.

\begin{figure}[ht]
\begin{center}
  \includegraphics[width=1.0\linewidth]{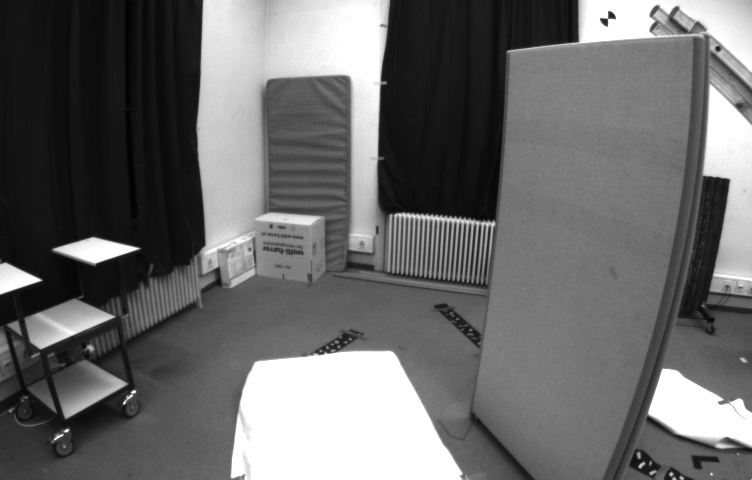}
      \includegraphics[width=1.0\linewidth]{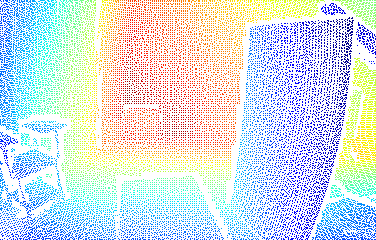}

\end{center}
  \caption{\small Illustration of a depth map (below) generated from the EuRoC point cloud of Vicon Room 2, by projecting onto the view of the RGB camera (above).}
\label{occlusion_ablation}
\end{figure}

\subsection{Odometry}
\begin{figure}[ht]
\begin{center}
      \hspace*{-0.05\linewidth}
      \includegraphics[width=1.1\linewidth]{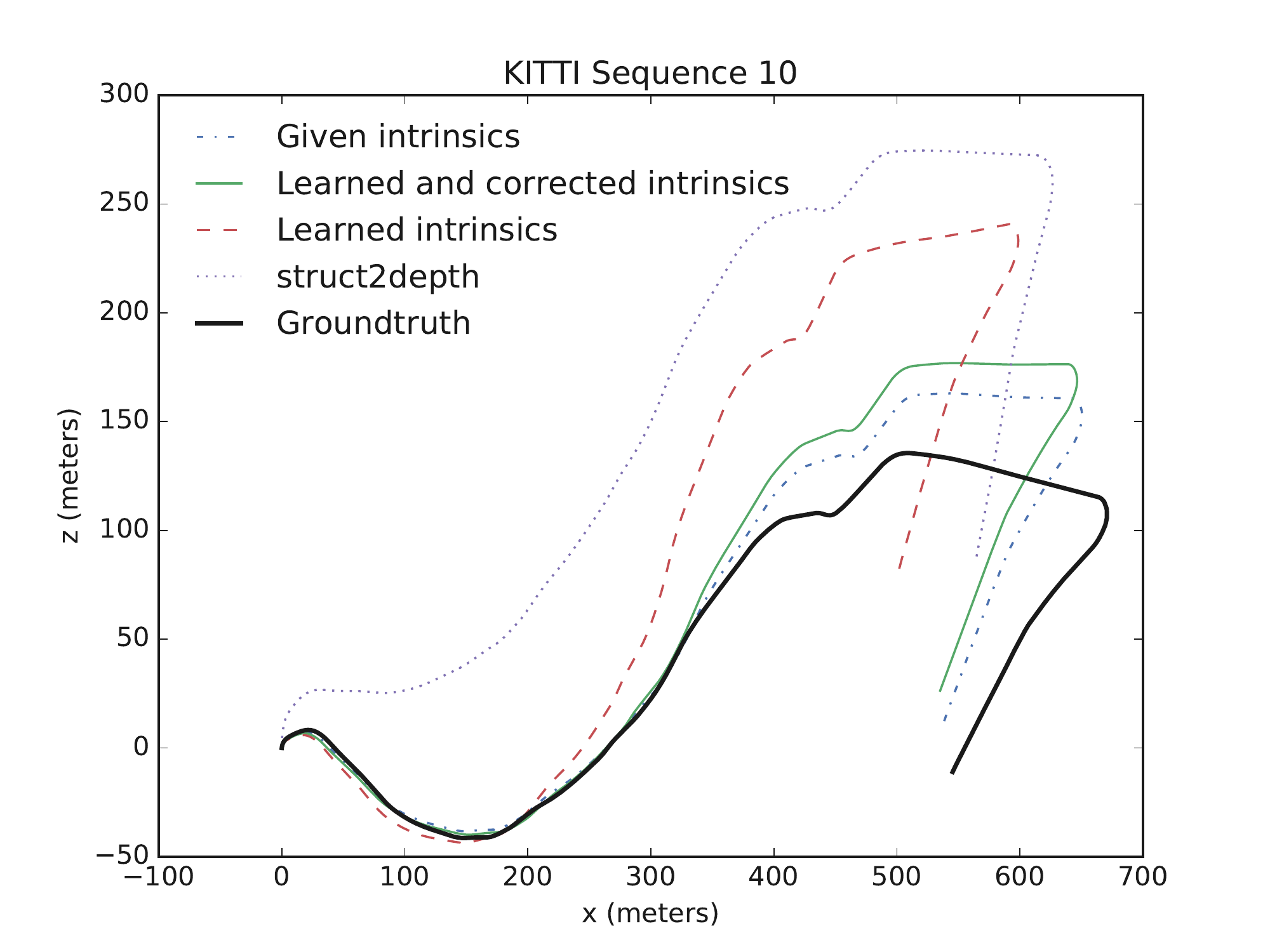}
\end{center}
  \caption{\small Predicted location on the KITTI odometry sequence 10, by a model trained with given, a model that learned the intrinsics, and the
  latter model with inference time correction applied. The groundtruth and the struct2depth \cite{casser2019struct2depth} results are displayed as well.}\label{odofig10}
\end{figure}

The KITTI Sequence 10 is shown in Figure \ref{odofig10}. Tables \ref{tab:ODOATE} and \ref{tab:odotrel} extend Table \ref{tab:odoboth} with more metrics.

\begin{table} [h!]
\small
  \centering 
  {
  \begin{tabular}{|l|c|c|}
  \hline
  Method & Seq.~09 & Seq.~10 \\
  \hline
  Zhou \cite{zhou2017unsupervised} & $0.021\pm0.017$ & $0.020\pm0.015$ \\
  Mahjourian \cite{mahjourian2018unsupervised} & $0.013\pm0.010$ & $0.012 \pm 0.011$ \\
  GeoNet \cite{yin2018geonet} & $ 0.012 \pm 0.007$ & $0.012 \pm 0.009$ \\
  Godard \cite{godard2018digging} & $0.023\pm0.013$ & $0.018\pm0.014$ \\
  Struct2depth \cite{casser2019struct2depth}& $0.011 \pm 0.006 $ & $0.011 \pm 0.010$ \\
  \hline
  Ours, with intrinsics: & &  \\
  Given & $\mathbf{0.009} \pm 0.015$ & $0.008 \pm 0.011$ \\
  Learned & $0.012 \pm 0.016$ & $0.010 \pm 0.010$ \\
  Learned \& corrected & $0.010 \pm 0.016$ & $\bf0.007 \pm 0.009$ \\
  \hline
  
  \end{tabular}
  }
  \vspace{1mm}
  \caption{\small 5-point Absolute Trajectory Error, (ATE) calculated following the procedure outlined in \cite{zhou2017unsupervised}. The three variants of our method are a model trained with given intrinsics, a model trained with learned intrinsics, and the latter model with test-time correction of the intrinsics. Our model outperforms prior art on Seq. 10, and in terms of the average error on Seq. 09. The larger variance of the error in Seq.~09 may be explained by the fact that we only use two frames to infer egomotion, whereas all others use 3 or more. In order to test the impact of these variations in the 5-point ATE metric, we calculated the relative translation error, which is more indicative of localization, and plotted the trajectories (Fig.~\ref{odofig} and Fig.~\ref{odofig10}). This table extends Table \ref{tab:odoboth} in the main paper.}
    \label{tab:ODOATE}
\end{table}

\begin{table} [ht]
\small
  \centering 
  {
  \begin{tabular}{|l||c|c||c|c|}
  \hline
    & \multicolumn{2}{c||}{{Seq.~09}} & \multicolumn{2}{c|}{{Seq.~10}} \\
        \hline 
  Method &$t_{rel}$  & $r_{rel}$ & $t_{rel}$  & $r_{rel}$ \\
  \hline
  Zhou \cite{zhou2017unsupervised} a la \cite{zhan2018unsupervised}&  17.8 & 6.78 & 37.9 & 17.8 \\
  Zhou \cite{zhou2017unsupervised} a la \cite{ShamwellOdometry}&  21.63 & 3.57 & 20.5 & 10.9\\
  Zhan \cite{zhan2018unsupervised}   & 11.9 & 3.60 & 12.6 & 3.43 \\ 
  Struct2depth \cite{casser2019struct2depth}& 10.2 & 2.64 & 29.0 & 4.28 \\
  \hline
  Ours, with intrinsics: & & & &\\
  Given &3.18 &0.586 &\textbf{5.38} &\textbf{1.03}\\
  Learned &7.47 &0.960 &13.2 & 3.09\\
  Learned \& corrected& \textbf{2.70} &\textbf{0.462} &6.87 &1.36 \\
  \hline
  \end{tabular}
  }
  \vspace{1mm}
  \caption{\small Average relative translation error ($t_{rel}$, in percents) and average relative rotation error ($r_{rel}$, degrees per 100 meters) calculated on the KITTI odometry sequences 09 and 10. The results for the method in Zhou et al. \cite{zhou2017unsupervised} were taken from two different evaluations \cite{ShamwellOdometry,zhan2018unsupervised}. The number for Struct2depth \cite{casser2019struct2depth} were evaluated using their published code and models. As in prior work, \cite{ShamwellOdometry,zhan2018unsupervised} the metrics are calculated starting after the first 100 meters.  This table extends Table \ref{tab:odoboth} in the main paper.}
    \label{tab:odotrel}
\end{table}

\end{document}